\documentstyle[aaai]{article}
\def\i#1{\hbox{\it #1\/}}
\def\no{\i{not}}
\def\lar{\leftarrow}
\def\beq{\begin{equation}}
\def\eeq#1{\label{#1}\end{equation}}
\def\ba{\begin{array}}
\def\ea{\end{array}}

\begin{document}

\title{\bf Fages' Theorem\\ and Answer Set Programming}
\author{Yuliya Babovich, Esra Erdem and Vladimir Lifschitz \\
        Department of Computer Sciences \\
        University of Texas at Austin \\
        Austin, TX 78712, USA \\
        Email: \{yuliya,esra,vl\}@cs.utexas.edu}
\date{today}

\maketitle
\begin{abstract}
We generalize a theorem by Fran\c{c}ois Fages that describes the
relationship between the completion semantics and the answer set
semantics for logic programs with negation as failure.  The study of
this relationship is important in connection with the emergence of
answer set programming.  Whenever the two semantics are equivalent,
answer sets can be computed by a satisfiability solver, and the
use of answer set solvers such as \hbox{\sc smodels} and
\hbox{\sc dlv} is unnecessary.  A logic programming representation of
the blocks world due to Ilkka Niemel\"a is discussed as an example.
\end{abstract}

\bibliographystyle{aaai}

\section{Introduction}

This note is about the relationship between the completion
semantics~\cite{cla78} and the answer set (``stable model'')
semantics~\cite{gel91b} for logic programs with negation as failure.
The study of this relationship is important in connection
with the emergence of answer set
programming~\cite{mar99,nie99,lif99c}.  Whenever the two semantics are
equivalent, answer sets can be computed by a satisfiability solver, and the
use of ``answer set solvers'' such as
\hbox{\sc smodels}\footnote{\tt http://saturn.hut.fi/pub/smodels/ .}
and \hbox{\sc dlv}\footnote{\tt http://www.dbai.tuwien.ac.at/proj/dlv/ .}
is unnecessary.


Consider a finite propositional (or grounded) program $\Pi$ without classical
negation, and a set $X$ of atoms.  If $X$ is an answer set for $\Pi$ then $X$,
viewed as a truth assignment, satisfies the completion of $\Pi$.  The
converse, generally, is not true.  For instance, the completion of
\beq
p\lar p
\eeq{Pi0}
is
$p\equiv p$.
This formula has two models $\emptyset$, $\{p\}$; the first is an answer set
for (\ref{Pi0}), but the second is not.  Fran\c{c}ois Fages~[\citeyear{fag94}]
defined a syntactic condition on logic programs
that implies the equivalence between the two
semantics---``positive-order-consistency,'' also called
``tightness''~\cite{lif96b}.  What he requires is the existence of
a function $\lambda$ from atoms to nonnegative integers (or, more
generally, ordinals) such that, for every rule
$$
A_0 \lar A_1, \ldots, A_m, \no\ A_{m+1}, \ldots, \no\ A_n
$$
in $\Pi$,
$$\lambda(A_1), \ldots, \lambda(A_m) < \lambda(A_0).$$
It is clear, for instance, that program (\ref{Pi0}) is not tight.
Fages proved that, for a tight program, every model of its completion is an
answer set.  Thus, for tight programs, the completion semantics and
the answer set semantics are equivalent.

Our generalization of Fages' theorem allows us to draw similar conclusions
for some programs that are not tight.  Here is one such program:
\beq
\ba l
p \lar \no\ q, \\
q \lar \no\ p, \\
p \lar p, r.
\ea
\eeq{Pi2}
It is not tight.  Nevertheless, each of the two models $\{p\}$, $\{q\}$ of
its completion
$$
\ba l
p \equiv \neg q \vee (p \wedge r), \\
q \equiv \neg p, \\
r \equiv \bot
\ea
$$
is an answer set for~(\ref{Pi2}).

The idea of this generalization is to make function $\lambda$ partial.
Instead of tight programs, we will consider programs that are
``tight on a set of literals.''

First we relate answer sets to a model-theoretic
counterpart of completion introduced
in~\cite{apt88}, called supportedness.  This allows us to make the theorem
applicable
to programs with both negation as failure and classical negation, and to
programs with infinitely many rules.\footnote{The familiar definition of
completion (see Appendix) is applicable to finite programs only, unless we
allow infinite disjunctions in completion formulas.}
Then a corollary about completion
is derived, and applied to
a logic programming representation of the blocks world due to
Ilkka Niemel\"a.  We show how the satisfiability
solver {\sc sato}~\cite{zha97} can be used to find answer sets for that
representation, and compare the performance of {\sc smodels} and {\sc sato}
on several benchmarks.

\section{Generalized Fages' Theorem} 

We define a {\sl rule} to be an expression of the form
\beq
\i{Head}\lar L_1,\dots,L_m,\no\ L_{m+1},\dots,\no\ L_n
\eeq{rule}
$(n\geq m\geq 0)$ where each $L_i$ is a literal (propositional atom possibly
preceded by classical negation $\neg$), and \i{Head} is a literal or the
symbol $\bot$.
A rule~(\ref{rule}) is called a {\sl fact} if $n=0$,
and a {\sl constraint} if $\i{Head}=\bot$.
A {\sl program} is a set of rules.
The familiar definitions of answer sets, closed sets and supported sets
for a program, as well as the definition of the completion of a program,
are reproduced in the appendix.

Instead of ``level mappings'' used by Fages, we
consider here {\em partial level mappings}---partial functions from literals
to ordinals.  A program $\Pi$ is {\em tight} on a set
$X$ of literals if there exists a partial level mapping $\lambda$ with
the domain $X$ such that, for every rule~(\ref{rule}) in $\Pi$, if
$\i{Head},L_1,\ldots,L_m\in X$ then
$$\lambda(L_1), \ldots, \lambda(L_m) < \lambda(\i{Head}).$$
(For the constraints in $\Pi$
this condition holds trivially, because the head of a
constraint is not a literal and thus cannot belong to $X$.)

\proclaim Theorem.
For any program $\Pi$ and any consistent set $X$ of literals such that
$\Pi$ is tight on $X$, $X$ is an answer set for~$\Pi$ iff $X$~is closed
under and supported by~$\Pi$.

The proof below is almost unchanged from the proof of Fages'
theorem given in~\cite[Section~7.4]{lif99b}.

\proclaim Lemma.
For any program $\Pi$ without negation as failure and any consistent set
$X$ of literals such that $\Pi$ is tight on $X$, if $X$~is closed under
and supported by~$\Pi$, then $X$~is an answer set for~$\Pi$.

\medskip\noindent{\bf Proof:}
We need to show that $X$~is minimal among the sets closed under~$\Pi$.
Assume that it is not.  Let $Y$ be a proper subset of~$X$ that is
closed under~$\Pi$, and let $\lambda$ be a partial level mapping
establishing that $\Pi$ is tight on $X$.  Take a literal~$L \in X\setminus Y$
such that $\lambda(L)$ is minimal.
Since $X$ is supported by $\Pi$, there is a rule
$$L\lar L_1,\dots,L_m$$
in $\Pi$ such that $L_1,\dots,L_m \in X$.  By the choice of $\lambda$,
$$\lambda(L_1),\dots,\lambda(L_m) < \lambda(L).$$
By the choice of~$L$, we can conclude that
$$L_1,\dots,L_m \in Y.$$
Consequently $Y$~is not closed under~$\Pi$, contrary to the choice of $Y$.

\medskip\noindent{\bf Proof of the Theorem:}
Left-to-right, the proof is straightforward.
Right-to-left: assume that $X$ is closed under and supported by~$\Pi$.
Then~$X$ is
closed under and supported by~$\Pi^X$.  Since $\Pi$ is tight on $X$, so is
$\Pi^X$.  Hence, by the lemma, $X$~is an answer set for~$\Pi^X$,
and consequently an answer set for~$\Pi$.
\medskip

In the special case when $\Pi$ is a
finite program without classical negation, a set of atoms satisfies the
completion of $\Pi$
iff it is closed under and supported by $\Pi$. We conclude:

\proclaim Corollary 1.
For any finite program $\Pi$ without classical negation and any set $X$ of
atoms such that $\Pi$ is tight on $X$, $X$ is an answer set for~$\Pi$ iff
$X$~satisfies the completion of~$\Pi$.

For instance, program~(\ref{Pi2}) is tight on the model $\{p\}$ of its
completion: take $\lambda(p) = 0$.
By Corollary 1, it follows that $\{p\}$ is an answer set for~(\ref{Pi2}).
In a similar way, the theorem shows that $\{q\}$ is an answer set also.

By $\i{pos}(\Pi)$ we denote the set of all literals that occur without
negation as failure at least once in the body of a rule of $\Pi$.

\proclaim Corollary 2.
For any program $\Pi$ and any consistent set $X$ of literals disjoint from
$\i{pos}(\Pi)$, $X$ is an answer set for~$\Pi$ iff $X$ is closed under and
supported by~$\Pi$.

\proclaim Corollary 3.
For any finite program $\Pi$ without classical negation and any set $X$ of
atoms disjoint from $pos(\Pi)$, $X$ is an answer set for~$\Pi$ iff $X$
satisfies the completion of~$\Pi$.

To derive Corollary~2 from the theorem, and Corollary~3 from Corollary~1,
take $\lambda(L)=0$ for every $L\in X$. 

Consider, for instance, the program
\beq
\ba l
p \lar \no\ q, \\
q \lar \no\ p, \\
r \lar r, \\
p \lar r.
\ea
\eeq{Pi1}
The completion of~(\ref{Pi1}) is
$$
\ba l
p \equiv \neg q \vee r, \\
q \equiv \neg p, \\
r \equiv r.
\ea
$$
The models of these formulas are $\{p\}$, $\{q\}$
and $\{p,r\}$.  The only literal occurring in the bodies of the rules
of~(\ref{Pi1}) without negation as failure is $r$.  In accordance with
Corollary~3, the models of the completion that do not contain $r$---sets
$\{p\}$ and $\{q\}$---are answer sets for~(\ref{Pi1}).

\section{Planning in the Blocks World} 

As a more interesting example, consider a logic programming encoding of
the blocks world due to Ilkka Niemel\"a.  The main part of the
encoding consists of the following schematic rules:

\begin{verbatim}

goal :- time(T), goal(T).
:- not goal. 

goal(T2) :- nextstate(T2,T1), goal(T1).

moveop(X,Y,T):-
   time(T), block(X), object(Y), X != Y, 
   on_something(X,T), available(Y,T), 
   not covered(X,T), not covered(Y,T),	
   not blocked_move(X,Y,T).

on(X,Y,T2) :- 
   block(X), object(Y), nextstate(T2,T1), 
   moveop(X,Y,T1).

on_something(X,T) :-
   block(X), object(Z), time(T), on(X,Z,T).

available(table,T) :- time(T).

available(X,T) :- 
   block(X), time(T), on_something(X,T).

covered(X,T) :-
   block(Z), block(X), time(T), on(Z,X,T).

on(X,Y,T2) :-
   nextstate(T2,T1), block(X), object(Y), 
   on(X,Y,T1), not moving(X,T1).
 
moving(X,T) :- time(T), block(X), object(Y), 
   moveop(X,Y,T).

blocked_move(X,Y,T):- 
   block(X), object(Y), time(T), goal(T).

blocked_move(X,Y,T) :- 
   time(T), block(X), object(Y), 
   not moveop(X,Y,T).

blocked_move(X,Y,T) :- 
   block(X), object(Y), object(Z), time(T), 
   moveop(X,Z,T), Y != Z.

blocked_move(X,Y,T) :- 
   block(X), object(Y), time(T), moving(Y,T).

blocked_move(X,Y,T) :- 
   block(X), block(Y), block(Z), time(T), 
   moveop(Z,Y,T), X != Z.

:- block(X), time(T), moveop(X,table,T), 
   on(X,table,T).

:- nextstate(T2,T1), block(X), object(Y),  
   moveop(X,Y,T1), moveop(X,table,T2).

nextstate(Y,X) :- time(X), time(Y), 
   Y = X + 1.

object(table).
object(X) :- block(X). 

\end{verbatim}

To solve a planning problem, we combine these rules with
\begin{enumerate}
\item[(i)] a set of facts defining {\tt time/1} as an initial segment of
nonnegative integers, for instance
\begin{verbatim}
  time(0).   time(1).   time(2).
\end{verbatim}
\item[(ii)] a set of facts defining {\tt block/1}, such as
\begin{verbatim}
  block(a).  block(b).  block(c).
\end{verbatim}
\item[(iii)] a set of facts encoding the initial state, such as
\begin{verbatim}
  on(a,b,0).  on(b,table,0).
\end{verbatim}
\item[(iv)] a rule that encodes the goal, such as
\begin{verbatim}
  goal(T) :- time(T), on(a,b,T), on(b,c,T).
\end{verbatim}
\end{enumerate}
The union is given as input to the ``intelligent grounding''
program~{\sc lparse}, and the result of grounding is passed on
to {\sc smodels}~\cite[Section~7]{nie99}.  The answer sets for the
program correspond to valid plans.

Concurrently executed actions are allowed in this formalization as long as
their effects are not in conflict, so that they can be arbitrarily interleaved.

The schematic rules above contain the variables
{\tt T}, {\tt T1}, {\tt T2}, {\tt X}, {\tt Y}, {\tt Z} that range over
the object constants occurring in the program, that is, over the
nonnegative integers that occur in the definition of {\tt time/1},
the names of blocks {\tt a}, {\tt b},$\dots$ that occur in the definition of
{\tt block/1}, and the object constant {\tt table}.

The expressions in the bodies of the schematic rules that contain {\tt =}
and {\tt !=} restrict the constants that are substituted for the variables
in the process of grounding.  For instance, we understand the schematic
rule
\begin{verbatim}
nextstate(Y,X) :- time(X), time(Y), 
   Y = X + 1.
\end{verbatim}
as an abbreviation for the set of all ground instances of
\begin{verbatim}
nextstate(Y,X) :- time(X), time(Y).
\end{verbatim}
in which {\tt X} and {\tt Y} are instantiated by a pair of consecutive
integers.  The schematic rule
\begin{verbatim}
blocked_move(X,Y,T) :- 
   block(X), object(Y), object(Z), time(T), 
   moveop(X,Z,T), Y != Z.
\end{verbatim}
stands for the set of all ground instances of
\begin{verbatim}
blocked_move(X,Y,T) :- 
   block(X), object(Y), object(Z), time(T),
   moveop(X,Z,T).
\end{verbatim}
in which {\tt Y} and {\tt Z} are instantiated by different object constants.

According to this understanding of variables and ``built-in predicates,''
Niemel\"a's schematic program, including rules (i)--(iv),
is an abbreviation for a finite program \i{BW} in the
sense defined above.

In the proposition below we assume that schematic rule (iv) has the form
\begin{verbatim}
   goal(T) :- time(T), ...
\end{verbatim}
where the dots stand for a list of schematic atoms with the predicate
symbol {\tt on} and the last argument {\tt T}.


\proclaim Proposition. Program~\i{BW} is tight on each of the models
of its completion.


\proclaim Lemma.  For any atom of the form {\tt nextstate(Y,X)} that
belongs to a model of the completion of program~\i{BW}, ${\tt Y}={\tt X}+1$.

\medskip\noindent{\bf Proof:}
The completion of~\i{BW} contains the formula
$$\tt{nextstate(Y,X)} \equiv \tt{false}$$
for all {\tt Y}, {\tt X} such that ${\tt Y}\neq{\tt X}+1$.

\medskip\noindent{\bf Proof of the Proposition.} 
Let $X$ be an answer set for \i{BW}.  By $T_{max}$ we denote the
largest argument of {\tt time/1} in its definition (i).
Consider the partial level mapping
$\lambda$ with domain $X$ defined as follows:
$$
\ba{l}
\lambda{\tt (time(T))              }= 0,                    \\ 
\lambda{\tt (block(X))             }= 0,                    \\ 
\lambda{\tt (object(X))            }= 1,                    \\ 
\lambda{\tt (nextstate(Y,X))       }= 1,                    \\ 
\lambda{\tt (covered(X,T))         }= 4\cdot {\tt T} + 3,   \\ 
\lambda{\tt (on\_something(X,T))   }= 4\cdot {\tt T} + 3,   \\ 
\lambda{\tt (available(X,T))       }= 4\cdot {\tt T} + 4,   \\ 
\lambda{\tt (moveop(X,Y,T))        }= 4\cdot {\tt T} + 5,   \\ 
\lambda{\tt (on(X,Y,T))            }= 4\cdot {\tt T} + 2,   \\ 
\lambda{\tt (moving(X,T))          }= 4\cdot {\tt T} + 6,   \\ 
\lambda{\tt (goal(T))              }= 4\cdot {\tt T} + 3,   \\ 
\lambda{\tt (blocked\_move(X,Y,T)) }= 4\cdot {\tt T} + 7,   \\ 
\lambda{\tt (goal)                 }= 4\cdot T_{max} + 4.
\ea
$$
This level mapping satisfies the inequality from the definition of a tight
program for every rule of \i{BW}; the lemma above
allows us to verify this assertion for the rules containing {\tt nextstate}
in the body.
\medskip

According to Corollary~1, we can conclude that the
answer sets for program \i{BW} can be equivalently characterized as the models
of the completion of \i{BW}.

\section{Answer Set Programming\\ with CCALC and SATO} 

The equivalence of the completion semantics to the answer set semantics
for program \i{BW} shows that it is not necessary to use an
answer set solver, such as {\sc smodels}, to compute answer sets for~\i{BW};
a satisfiability solver can be used instead.  Planning by giving the
completion of \i{BW} as input to a satisfiability solver is a
form of answer set programming and, at the same time, a special case of
satisfiability planning~\cite{kau92}.

The Causal Calculator, or
{\sc ccalc}\footnote{\tt http://www.cs.utexas.edu/users/tag/ccalc/ .}, 
is a system that is capable, among other things, of grounding and completing a
schematic logic program, clausifying the completion, and calling
a satisfiability solver (for instance, {\sc sato}) to find a model.
We have conducted a series of experiments aimed at comparing the run times of
{\sc sato}, when its input is generated from \i{BW} by {\sc ccalc}, with the
run times of {\sc smodels}, when its input is generated from \i{BW}
by~{\sc lparse}.

Because the built-in arithmetic of {\sc ccalc} is somewhat different from
that of {\sc lparse}, we had to modify \i{BW} slightly.  Our {\sc ccalc}
input file uses variables of sorts {\tt object}, {\tt block} and {\tt time}
instead of the unary predicates with these names.  The rules of \i{BW}
that contain those predicates in their bodies are modified accordingly.
For instance, rule
\begin{verbatim}
on_something(X,T) :-
    block(X), object(Z), time(T), on(X,Z,T).
\end{verbatim}
turns into
\begin{verbatim}
on_something(B1,T) :- on(B1,O2,T).
\end{verbatim}
The macro expansion facility of {\sc ccalc} expands 
\begin{verbatim}
   nextstate(T2,T1)
\end{verbatim}
into the expression 
\begin{verbatim}
   T2 is T1 + 1
\end{verbatim}
that contains Prolog's built-in {\tt is}.

\begin{figure}
$$
\begin{tabular}{lrrrr}
Problem &Blocks    &Steps    &Run time            &Run time \\
        &          &         &of                  &of  \\  
        &          &         &{\sc smodels}       &{\sc sato}   \\
\hline
large.c &15  &7   &9.86            &1.82\\
        &    &8   &31.25           &2.16\\
\hline
large.d &17  &8   &18.25           &2.96\\
        &    &9   &62.48           &4.14\\
\hline
large.e &19  &9   &27.31           &5.40\\
        &    &10  &101.4$\;\,$     &7.16\\
\hline
\end{tabular}
$$
\caption{Planning with \i{BW}: {\sc sato} vs.~{\sc smodels}}
\end{figure}

Figure~1 shows the run times of {\sc smodels} (Version 2.24 and {\sc sato}
(Version 3.1.2) in seconds,
measured using the Unix {\tt time} command, on the benchmarks
from~\cite[Section~9, Table~3]{nie99}.  For each problem, one of the two
entries corresponds to the largest number of steps for which the
problem is not solvable, and the other to the smallest number of steps for
which a solution exists.
The experiments were performed on
an UltraSPARC with 124 MB main memory and a 167 MHz CPU.

The numbers in
Figure~1 are ``search times''---the grounding and completion times are
not included.  The computation involved in grounding and completion does
not depend on the initial state or the goal of the planning problem and,
in this sense,
can be viewed as ``preprocessing.''  {\sc lparse} performs
grounding more efficiently than {\sc ccalc}, partly because the former is
written in
C$++$ and the latter in Prolog.  The last benchmark in Figure~1 was grounded
by {\sc lparse} (Version 0.99.41) in 16 seconds; {\sc ccalc} (Version 1.23)
spent 50
seconds in grounding and about the same amount of time forming the completion.
But the size of the grounded program is approximately the same in both cases:
{\sc lparse} generated 191621 rules containing 13422 atoms, and
{\sc ccalc} generated 200661 rules containing 13410 atoms.

\section{Discussion} 

Fages' theorem, and its generalization proved in this note,
allow us to compute answer sets for some programs by completing them and
then calling a satisfiability solver.  We showed that this method can be
applied, for instance, to the representation of
the blocks world proposed in~\cite{nie99}.
This example shows that satisfiability solvers may serve as useful
computational tools in answer set programming.

There are cases, on the other hand, when the completion method is not
applicable.  Consider computing Hamiltonian cycles in a directed
graph~\cite{mar99}.  We combine the rules

\begin{verbatim}
in(U,V) :- edge(U,V), not out(U,V).
out(U,V) :- edge(U,V), not in(U,V).

:- in(U,V), in(U,W), V != W.
:- in(U,W), in(V,W), U != V.

reachable(V) :- in(v0,V).
reachable(V) :- reachable(U), in(U,V).

:- vertex(U), not reachable(U).
\end{verbatim}

\noindent
with a set of facts defining the vertices and edges of the graph; {\tt v0} is
assumed to be one of the vertices.  The answer sets for the resulting
program correspond to the Hamiltonian cycles.  Generally, the completion of
the program has models different from its answer sets.  Take, for instance,
the graph consisting of two disjoint loops:
\begin{verbatim}
   vertex(v0).  vertex(v1).
   edge(v0,v0).  edge(v1,v1).
\end{verbatim}
This graph has no Hamiltonian cycles, and, accordingly, the corresponding
program has no answer sets.  But the set 
\begin{verbatim}
   vertex(v0), vertex(v1), edge(v0,v0), 
   edge(v1,v1), in(v0,v0), in(v1,v1), 
   reachable(v0), reachable(v1)
\end{verbatim}
is a model of the program's completion.

\section*{Acknowledgements}

We are grateful to Victor Marek, Emilio Remolina,
Mirek Truszczy\'nski and Hudson Turner, and to the anonymous referees,
for comments and criticisms.
This work was partially supported by National
Science Foundation under grant IIS-9732744.


\section*{Appendix: Definitions}

The notion of an answer set is defined first for programs whose rules do not
contain negation as failure.
Let $\Pi$ be such a program, and let $X$ be a consistent set of literals.
We say that $X$ is {\sl closed} under $\Pi$ if, for every rule
$$\i{Head}\lar\i{Body}$$
in $\Pi$, $\i{Head}\in X$ whenever
$\i{Body}\subseteq X$.  (For a constraint, this condition means that
the body is not contained in $X$.)  We say that $X$ is an {\sl answer set}
for $\Pi$ if $X$ is minimal among the sets closed under $\Pi$ w.r.t.~set
inclusion.  It is clear that a program without negation as failure can have
at most one answer set.

To extend this definition to arbitrary programs, take any program $\Pi$, and
let $X$ be a consistent set of literals.  The {\sl reduct} $\Pi^X$
of $\Pi$ relative to $X$ is the set of rules
$$
\i{Head}\lar L_1,\dots,L_m
$$
for all rules~(\ref{rule}) in $\Pi$ such that $L_{m+1},\dots,L_n\not\in X$.
Thus $\Pi^X$ is a program without negation as failure.  We say that $X$
is an {\sl answer set} for~$\Pi$ if $X$~is an answer set for~$\Pi^X$.

A set $X$ of literals is
{\sl closed under} $\Pi$ if, for every rule~(\ref{rule})
in $\Pi$, ${\i{Head} \in X}$ whenever ${L_1,\ldots,L_m\in X}$
and ${L_{m+1},\dots,L_n\not\in X}$.
We say that $X$~is {\sl supported by}~$\Pi$ if,
for every ${L \in X}$, there is a rule~(\ref{rule}) in~$\Pi$
such that~${\i{Head} = L}$, ${L_1,\ldots,L_m\in X}$ and
${L_{m+1},\dots,L_n\not\in X}$.

Let $\Pi$ be a finite program without classical negation. If $H$ is an atom
or the symbol $\bot$, by $\i{Comp}(\Pi,H)$ we denote the formula
$$
H \equiv \bigvee
 (A_1\wedge\cdots\wedge A_m\wedge\neg A_{m+1}\wedge\cdots\wedge\neg A_n)
$$
where the disjunction extends over all rules
$$
H\lar A_1,\dots,A_m,\no\ A_{m+1},\dots,\no\ A_n
$$
in $\Pi$ with the head $H$.  The {\sl completion} of $\Pi$ is set of formulas
$\i{Comp}(\Pi,H)$ for all $H$.

\end{document}